\pdfoutput=1   

\documentclass[11pt]{article}

\usepackage[margin=1in]{geometry}
\usepackage[numbers,compress]{natbib}

\usepackage[utf8]{inputenc}
\usepackage[T1]{fontenc}
\usepackage{lmodern}   
\usepackage{url}
\usepackage{booktabs}
\usepackage{amsmath,amssymb,amsfonts}
\usepackage{nicefrac}
\usepackage{microtype}
\usepackage{xcolor}
\usepackage{algorithmic}
\usepackage{algorithm}
\usepackage{graphicx}
\usepackage{textcomp}
\usepackage{float}

\usepackage{hyperref}
\hypersetup{
  colorlinks=true,
  linkcolor=blue!55!black,
  citecolor=blue!55!black,
  urlcolor=blue!55!black,
}

\title{Enhancing Bayesian Optimization and Active Learning\\ Through Kernel Diversity\thanks{H. Zhang, H. Xiang and Q. Lu are supported by NSF \#2340049.}}

\author{
Heng Zhang\textsuperscript{1} \quad
Haotian Xiang\textsuperscript{1} \quad
Konstantinos D. Polyzos\textsuperscript{2} \quad
Tara Javidi\textsuperscript{2} \quad 
Qin Lu\textsuperscript{1} \quad
\\[3pt]
\textsuperscript{1}University of Georgia, Athens, GA, USA\\
\textsuperscript{2}University of California San Diego, La Jolla, CA, USA
\\[3pt]
\texttt{hz39726@uga.edu} \quad
\texttt{haotian.xiang@uga.edu} \quad
\texttt{kpolyzos@ucsd.edu} 
\\
\texttt{tjavidi@ucsd.edu}
\texttt{qin.lu@uga.edu} \quad
}

\date{}

\begin{document}

\maketitle

\begin{abstract}
Hyperparameter selection remains a key challenge in Bayesian optimization (BO) and Bayesian active learning (AL), as model misspecification can lead to suboptimal performance, while more accurate fully Bayesian treatments typically rely on computationally expensive MCMC sampling.  This paper proposes a unified framework, KENDO (Kernel ENsemble Disagreement-aware Operator), that integrates Ensemble Gaussian Processes (EGP) with disagreement-aware acquisition strategies. The central idea is to replace hyperparameter sampling with a kernel ensemble and adaptive Bayesian weighting, combined with disagreement-aware acquisition strategies. Within this unified framework, we instantiate KENDO-BO for BO and KENDO-AL for Bayesian AL, demonstrating that both arise from a common self-correcting mechanism with task-specific acquisition objectives. We further extend the approach to multi-objective optimization via random scalarization that preserves the single-optimizer conditioning structure. Thorough numerical tests on synthetic and real-world benchmarks across single-objective optimization, multi-objective optimization, and active learning demonstrate that (i) KENDO-BO achieves competitive or superior optimization performance compared to state-of-the-art methods while reducing computational overhead by up to $5\times$ and (ii) KENDO-AL achieves superior predictive calibration over MCMC-based active learning baselines with up to $27\times$ speedup.
\end{abstract}

\section{Introduction}
\label{sec:intro}
{Bayesian optimization (BO) and active learning (BAL) have become standard approaches for optimizing expensive black-box functions \citep{frazier2018tutorial, shahriari2016taking} and for learning unknown functions in an uncertainty-aware, label-efficient fashion respectively}. By constructing a probabilistic surrogate model, typically a Gaussian process (GP), {both paradigms} quantify uncertainty to guide efficient sequential sampling. However, the performance of GP-based BO and AL critically depends on hyperparameter selection, including kernel choice and lengthscale parameters \citep{snoek2012practical}.
Traditional approaches handle hyperparameters through point estimation via marginal likelihood maximization \citep{rasmussen2006gaussian} or full Bayesian treatment via MCMC sampling \citep{snoek2012practical}. Point estimation is computationally efficient but vulnerable to model misspecification and overfitting, especially in early optimization stages. Conversely, MCMC-based fully Bayesian methods account for hyperparameter uncertainty but incur prohibitive computational costs, particularly when sampling from multimodal or high-dimensional posteriors \citep{springenberg2016bayesian}.

\begin{figure}[t]
    \centering
    \includegraphics[width=\textwidth]{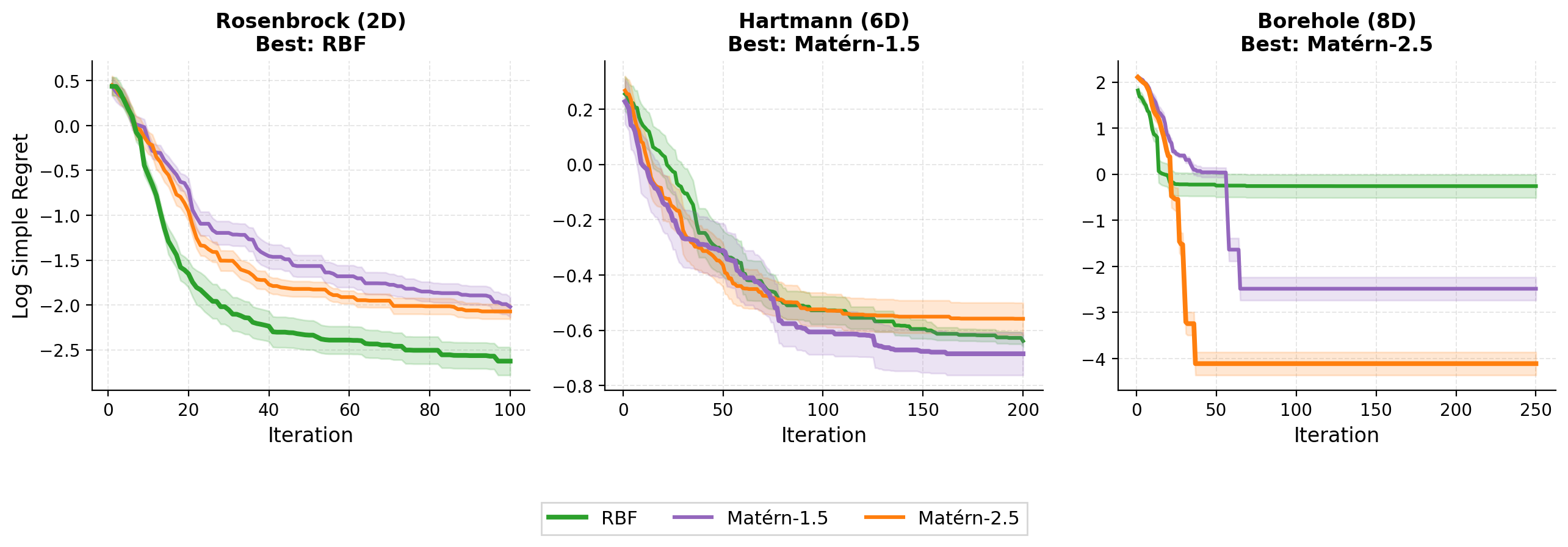}
    \caption{Single-kernel BO on three benchmarks for PES method, reported as $\log_{10}$ simple regret. No single kernel is universally optimal.}
    \label{fig:kernel_motivation}
\end{figure}

{Beyond hyperparameter tuning, the choice of kernel family itself is a critical source of misspecification. {For example, in the BO paradigm,} as Figure~\ref{fig:kernel_motivation} illustrates, no single kernel is universally optimal for BO: the best-performing kernel differs across problems, RBF on Rosenbrock (2D), Mat\'{e}rn-1.5 on Hartmann (6D), and Mat\'{e}rn-2.5 on Borehole (8D) in $\log_{10}$ simple regret, motivating an ensemble that adaptively weights multiple kernels based on observed data.}
Recent work have explored ensemble-based alternatives. Self-Correcting Bayesian Optimization (SCoreBO) \citep{scorebo} introduces an acquisition function that conditions on sampled optimizers to balance hyperparameter learning with optimization, but still relies on hyperparameter sampling. Ensemble Gaussian Processes (EGP) \citep{egp, lu2023surrogate, polyzos2023bayesian} maintain multiple GPs with different kernels and adaptively weight them via Bayesian model averaging, avoiding sampling entirely. {Nonetheless, these approaches adaptively select a single GP model and couple it with standard acquisition strategies, thereby overlooking the benefits from interaction and disagreement among GP models during the acquisition phase. To the best of our knowledge,} using EGPs as surrogate models in place of a fully Bayesian treatment, and integrating them with appropriate related acquisition strategies, remains largely unexplored.

\textbf{Our Contributions.} We propose KENDO (Kernel ENsemble Disagreement-aware Operator), which unifies the strengths of ensemble modeling and self-correcting optimization:
\begin{itemize}
    \item We replace SCoreBO's hyperparameter sampling with EGP's explicit kernel ensemble, eliminating MCMC overhead while preserving uncertainty quantification through kernel diversity and adaptive weighting.
    \item We derive a computationally efficient acquisition function that measures disagreement between marginal and optimizer-conditioned ensemble posteriors, balancing exploration in promising regions with model selection.
    \item We extend the framework to multi-objective optimization via random scalarization, which preserves the single-optimizer conditioning structure of SCoreBO while covering the Pareto front through diverse scalarization directions, avoiding the computational burden of multi-objective solvers.
    \item We introduce KENDO-AL, demonstrating that the ensemble mechanism generalizes beyond optimization to Bayesian active learning, achieving competitive results over MCMC-based approaches.
    \item Through extensive experiments and ablation studies, we demonstrate that the ensemble mechanism achieves substantial computational savings (up to $5\times$ for BO and $27\times$ for active learning) without sacrificing optimization performance.
\end{itemize}

\section{Related Work}
\label{sec:related}
\subsection{Gaussian Processes and Hyperparameter Learning}

Gaussian processes provide a flexible non-parametric framework for probabilistic regression \citep{rasmussen2006gaussian}, in which the choice of kernel function and its hyperparameters fundamentally determines the inductive bias of the surrogate model. Standard practice optimizes these hyperparameters by maximizing the marginal likelihood \citep{rasmussen2006gaussian}, but the resulting point estimate neglects uncertainty and can produce overconfident predictions \citep{snoek2012practical}. Fully Bayesian alternatives integrate over hyperparameter posteriors via MCMC \citep{snoek2012practical} or variational inference \citep{hensman2015scalable,titsias2009variational}, and scalable variants such as slice sampling \citep{murray2010slice} and Hamiltonian Monte Carlo \citep{hoffman2014nuts} improve robustness at the cost of substantial computational overhead where the surrogate must be refitted at every iteration. A complementary line of work addresses model uncertainty through ensembles rather than integration. Multi-kernel learning \citep{duvenaud2013structure} constructs expressive composite kernels and deep kernel learning \citep{wilson2016deep} parameterizes kernels via neural networks, while Ensemble Gaussian Processes \citep{egp} explicitly maintain a weighted collection of GPs with different kernels and update the weights via Bayesian model averaging based on predictive likelihood. This approach has shown promise but has not been integrated with modern BO or active learning acquisition functions that leverage optimizer information.

\subsection{Acquisition Functions for Bayesian Optimization and Active Learning}

A unifying theme connecting modern acquisition functions for both Bayesian optimization and Bayesian active learning is the use of \emph{disagreement} as the exploration signal, where queries are placed at points on which competing posterior beliefs disagree most strongly about the prediction. SCoreBO \citep{scorebo} instantiates this idea in BO by conditioning on sampled optimizers and computing the Hellinger distance between the marginal posterior and each optimizer-conditional posterior in promising regions, although it inherits the heavy computational cost of hyperparameter sampling. Other BO methods follow alternative routes that do not exploit disagreement, including GP-UCB with fixed confidence bounds \citep{srinivas2010gaussian}, entropy-search variants that integrate over functions \citep{hennig2012entropy,hernandez2014predictive,wang2017mes}, and meta-learning approaches that transfer hyperparameter knowledge across related tasks \citep{feurer2015initializing}. The same disagreement principle has long driven Bayesian active learning (BAL), where the goal shifts from finding optima to minimizing global prediction error \citep{settles2009active}. Bayesian Active Learning by Disagreement (BALD) \citep{houlsby2011bayesian,kirsch2019batchbald} maximizes the mutual information between predictions and hyperparameters, Bayesian Query-by-Committee (BQBC) \citep{riis2022bayesian} measures variance in posterior means across hyperparameter samples, and Statistical distance-based Active Learning (SAL) \citep{scorebo} generalizes BQBC using Hellinger and Wasserstein distances to capture the full distributional disagreement induced by hyperparameter uncertainty. SAL thus plays the same role in active learning as SCoreBO does in optimization, and both rely on MCMC sampling to quantify hyperparameter uncertainty, an overhead that directly motivates our ensemble-based alternative.

\subsection{Multi-Objective Bayesian Optimization}

Multi-objective BO (MOBO) aims to approximate the Pareto front of multiple conflicting objectives \citep{knowles2006parego}, and standard approaches include expected hypervolume improvement \citep{emmerich2006single,daulton2021qnehvi}, $\epsilon$-indicator-based methods \citep{zitzler2004ibea}, and information-theoretic acquisitions \citep{hernandez2016predictive,tu2022jesmo,belakaria2019mesmo}. More recent work explores scalarization strategies \citep{paria2020flexible} and uncertainty-aware acquisition functions \citep{daulton2020differentiable}, yet hyperparameter learning in multi-objective settings remains largely underexplored, as most existing methods either fix hyperparameters in advance or tune them independently for each objective without accounting for the coupled uncertainty across objectives.

\section{Preliminaries}
\label{sec:prelim}
Both BO and BAL are sequential decision-making paradigms that deal with a {\it black-box} function $f$, which has no analytic expression and is also expensive to evaluate. While BO aims to seek the optimizer of $f$, namely,
\begin{align}
    {\bf x}_* = \underset{{\bf x}\in {\cal X}}{\arg\max}\ \ \ f({\bf x})
\end{align}
BAL focuses on learning a good mapping for the target function $f$ for every ${\bf x}\in {\cal X}$ .

Specifically, BO and BAL rely on a probabilistic surrogate model that guides the judicious selection of query points sequentially, which are implemented iteratively in the two steps, \textbf{s1)}  obtain  $p(f({\bf x})|\mathcal{D}_t)$ ($\mathcal{D}_t:=\{({\bf x}_\tau, y_\tau) \}_{\tau = 1}^t$) based on a chosen surrogate model; and \textbf{s2)} select ${\bf x}_{t+1}\! =\! {\arg\max}_{{\bf x}\in {\cal X} }  \ \alpha ({\bf x}|\mathcal{D}_t)$,  where the acquisition function (AF) $\alpha$, usually available in closed form, is designed based on $p(\! f(\mathbf{x})|\mathcal{D}_t)$. For BO, the AF seeks to strike a balance between {\it exploration} and {\it exploitation}, while the AF in BAL is designed mainly for {\it exploration}.

\subsection{GP-based surrogate models}

Capable of learning function mappings with well-calibrated uncertainty values, GPs are the de facto choice for surrogate models in BO and BAL~\citep{rasmussen2006gaussian,frazier2018tutorial}. In the GP context, the learning function $f$ is assumed to be drawn from a GP prior, $f({\bf x}) \sim {\cal GP}(0, \kappa_{\boldsymbol{\theta}}({\bf x}, {\bf x}{'}))$, where $\kappa_{\boldsymbol{\theta}}$ is the positive-definite {\it kernel} function with learnable hyperparameters $\boldsymbol{\theta}$ that measures  pairwise similarity between any two {distinct points ${\bf x}$, and ${\bf x}'$ and is used to capture the covariance between}
function values $f({\bf x})$, $f({\bf x}{'})$. Given $\mathcal{D}_t$, a typical process is to  maximize the so-termed marginal likelihood $p({\cal D}_t; \boldsymbol{\theta})$ to obtain a {\it point} estimate of $\boldsymbol{\theta}$. Going beyond the point estimation of the kernel hyperparameters, the {\it fully Bayesian} GP views $\boldsymbol{\theta}$ as random and maintains a posterior pdf for $\boldsymbol{\theta}$ via Bayes' rule as: $p(\boldsymbol{\theta}|{\cal D}_t)\propto p(\boldsymbol{\theta}) p({\cal D}_t; \boldsymbol{\theta})$. Although accounting for overfitting, such a fully Bayesian treatment entails computationally prohibitive Markov Chain Monte Carlo (MCMC) sampling.

The previous discussion adheres to GPs with a {\it pre-selected} kernel type. In practice though, this is a nontrivial design choice as different tasks may require different kernel functions. To automate this design, the ensemble (E) GP framework \citep{egp}, given a prescribed kernel dictionary $\mathcal{K} := \{\kappa_1, \ldots, \kappa_M\}$, learns the function via a Gaussian mixture prior as~\cite{polyzos2024weighted,polyzos2021ensemble}
\begin{equation}
f({\bf x}) \sim \sum_{m=1}^M w_0^m \, {\cal GP}(0, \kappa_m({\bf x}, {\bf x}')), \quad \sum_{m=1}^M w_0^m = 1
\end{equation}
{where $w_0^m$ is the weight corresponding to the $m$th GP model using kernel $\kappa^m$}. After observing $\mathcal{D}_t$, the function posterior pdf is a GP mixture
\begin{equation}
p(f({\bf x})\mid \mathcal{D}_t) = \sum_{m=1}^M w_t^m \, \mathcal{N}(f({\bf x}); \mu_{m,t}({\bf x}), \sigma_{m,t}^2({\bf x}))
\label{eq:gp-mixture}
\end{equation}
where $p(f(\mathbf{x})|m, \mathcal{D}_t) := \mathcal{N}(f({\bf x}); \mu_{m,t}({\bf x}), \sigma_{m,t}^2({\bf x}))$ is the posterior of GP model $m$ with mean and variance given in closed-form \citep{rasmussen2006gaussian}. The per-kernel weight $w_t^m = \mathbb{P}(m|{\cal D}_t)$ is proportional to the marginal likelihood
\begin{equation}
w^m_{t} \propto p({\cal D}_t|m; \hat{\boldsymbol{\theta}}_m)
\label{eq:weight-update}
\end{equation}
with $\hat{\boldsymbol{\theta}}_m$ being the estimated hyperparameters for kernel $\kappa^m$ by maximizing the marginal likelihood.

\subsection{Disagreement-based AF}
Based on different design rules, a number of AFs have been designed for BO and BAL tailored to the chosen surrogate model. Most recently, leveraging the hyperparameter-induced disagreement conditioned on optimizer samples, a novel AF, termed as SCoreBO~\citep{scorebo}, is devised for the fully Bayesian GP model, defined as
\begin{equation}
\alpha_{\text{SC}}(\mathbf{x}) = \mathbb{E}_{\boldsymbol{\theta}, \ast} [d(p(y_\mathbf{x} \mid \mathcal{D}), p(y_\mathbf{x} \mid \boldsymbol{\theta}, \ast, \mathcal{D}))]
\end{equation}
where $\boldsymbol{\theta}$ are hyperparameters sampled from $p(\boldsymbol{\theta} \mid \mathcal{D})$, $\ast = (\mathbf{x}^*, f^*)$ is a sampled optimizer, and $d(\cdot, \cdot)$ is the Hellinger distance. This balances exploration in high-potential regions with hyperparameter learning. Unlike TS or EI that solely exploit the current posterior, $\alpha_{\text{SC}}$ also rewards queries that reduce hyperparameter uncertainty, which is critical when the surrogate is misspecified in early iterations.

\section{Exploiting Kernel Diversity to Enhance BO and BAL}
\label{sec:method}
While SCoreBO relies on the diversity of kernel hyperparameters, it still presumes a single pre-selected kernel type and uses computationally heavy MCMC sampling. {To further account for the effect of kernel family selection and to eliminate the computational burden of MCMC, we propose to replace hyperparameter sampling with an explicit kernel ensemble. The core idea lies on the  disagreement across structurally different kernels instead of the disagreement across hyperparameter samples of a single kernel. We develop this idea in three stages: first establishing the ensemble-based surrogate (Section \ref{from hyperparameter sampling to kernel ensembles}), then deriving the optimizer-conditioned acquisition function (Section \ref{optimizer-conditioned af}) {for (multi-objective) black-box optimization}, and finally extending to active learning (Section \ref{extensions to al}).}

\subsection{From Hyperparameter Sampling to Kernel Ensembles}
\label{from hyperparameter sampling to kernel ensembles}
The central idea of KENDO is to replace SCoreBO's continuous hyperparameter sampling with EGP's discrete kernel ensemble. Each GP $m \in \{1,\ldots,M\}$ relies on a fixed kernel $\kappa_m$ whose hyperparameters are fitted via marginal likelihood maximization, together with a dynamic weight $w^m_t$ updated via \eqref{eq:weight-update}. This substitution offers two advantages: (i) it eliminates MCMC sampling overhead; and (ii) it captures a qualitatively different and more consequential axis of model uncertainty, namely the choice of kernel family, {without relying on a specific pre-defined kernel form}. The Bayesian weight update requires only $M$ predictive likelihood evaluations per iteration, compared to the $\mathcal{O}(100)$--$\mathcal{O}(1000)$ posterior samples typically needed for MCMC convergence.

To make the ensemble tractable so as to be readily used as input to acquisition functions, we approximate the Gaussian mixture posterior \eqref{eq:gp-mixture} via moment matching as
\begin{align}
\bar{\mu}_t(\mathbf{x}) &= \sum_{m=1}^M w^m_t \, \mu_{m,t}(\mathbf{x}) \label{eq:mean}\\
\bar{\sigma}^2_t(\mathbf{x}) &= \sum_{m=1}^M w^m_t \left[ \sigma_{m,t}^2({\bf x}) + (\mu_{m,t}(\mathbf{x}) - \bar{\mu}_t(\mathbf{x}))^2 \right] \label{eq:var}
\end{align}
{where $\bar{\mu}_t(\mathbf{x})$ is the ensemble mean for any $\mathbf{x}$ and $\sigma_{m,t}({\bf x})$ is the associated variance.} The variance expression decomposes into within-GP uncertainty (first term) and between-GP disagreement (second term), providing a single Gaussian summary that retains ensemble-level information.

\subsection{Optimizer-Conditioned AF}
\label{optimizer-conditioned af}
{With a tractable Gaussian summary of the ensemble posterior, we will subsequently utilize this ensemble to decide where to query next. We adapt the key idea of conditioning on sampled optimizers and disagreement across hyperparameter samples of a single kernel, to the ensemble setting. The intuition is that if knowing the location of the optimum would cause different kernels to revise their predictions differently at $\mathbf{x}$, then querying $\mathbf{x}$ is informative for both optimization and model selection.} We define the acquisition function of KENDO-BO. For each kernel $m$ and sample index $n$, we draw a function realization $f_{m,n}$ from the posterior of GP $m$, locate its optimizer $\mathbf{x}^*_{m,n} = \arg\max_x f_{m,n}(\mathbf{x})$ with value $f^*_{m,n}$, and condition GP $m$ on $(\mathbf{x}^*_{m,n}, f^*_{m,n})$ to obtain a conditional posterior $\mathcal{N}(\mu^{\ast_{m,n}}_{m,t}(x), \sigma^{2,\ast_{m,n}}_{m,t}(\mathbf{x}))$. The acquisition function then aggregates the disagreement between the marginal ensemble posterior and each conditional posterior, yielding (see Appendix~\ref{app:derivation} for the full derivation)
\begin{equation}
\alpha(\mathbf{x}) = \sum_{m=1}^M \sum_{n=1}^N \frac{w^m_t}{N} \cdot d_{m,n}(\mathbf{x}) \label{eq:kendo-bo}
\end{equation}
where $d_{m,n}(\mathbf{x})$ is the Hellinger distance between the moment-matched marginal (\eqref{eq:mean}--\eqref{eq:var}) and the conditional posterior of kernel $m$ given optimizer sample $n$. For two Gaussians, this admits a closed-form expression
\begin{equation}
d_{m,n}(\mathbf{x}) = 1 - \sqrt{\frac{2\bar{\sigma}_t(\mathbf{x}) \sigma^{\ast_{m,n}}_{m,t}(\mathbf{x})}{\bar{\sigma}^2_t(\mathbf{x}) + \sigma^{2,\ast_{m,n}}_{m,t}(\mathbf{x})}} \exp\!\left[ -\frac{(\bar{\mu}_t(\mathbf{x}) - \mu^{\ast_{m,n}}_{m,t}(\mathbf{x}))^2}{4(\bar{\sigma}^2_t(\mathbf{x}) + \sigma^{2,\ast_{m,n}}_{m,t}(\mathbf{x}))} \right] \label{eq:hellinger-bo}
\end{equation}

Intuitively, $\alpha(\mathbf{x})$ is large at locations where knowing the optimizer would substantially revise the ensemble's prediction. Algorithm~\ref{alg:egp-unified} summarizes the full procedure.

\begin{algorithm}[t]
\caption{KENDO-BO~/~KENDO-AL}
\label{alg:egp-unified}
\small
\begin{algorithmic}[1]
\REQUIRE Kernel dict.\ $\mathcal{K}$, data $\mathcal{D}_{0}$, budget $T$ \hfill {\color{gray}\textit{// For BO: \#optimizers $N$;\ \ for AL: validation set $\mathcal{V}$}}
\FOR{$m \in \mathcal{K}$}
    \STATE Fit $\alpha_m$ via $\max p(\mathcal{D}_{0}|i=m,\alpha)$; \quad Init $w_0^m = 1/M$
\ENDFOR
\FOR{$t = 0, \ldots, T-1$}
    \STATE Compute $\bar{\mu}_t(\cdot), \bar{\sigma}^2_t(\cdot)$ via \eqref{eq:mean}--\eqref{eq:var}
    \STATE {\color{gray}\textit{// BO branch: sample optimizers and condition}}
    \FOR{$m \in \mathcal{K}$; $n = 1, \ldots, N$}
        \STATE Sample $f_{m,n} \sim p(f | m, \mathcal{D}_t)$; \quad Find $x^*_{m,n} = \arg\max f_{m,n}(\mathbf{x})$
        \STATE Condition GP $m$ on $(x^*_{m,n}, f^*_{m,n})$ to get $\mu^{*}_{m,n}, \sigma^{2,*}_{m,n}$
    \ENDFOR
    \STATE {\color{gray}\textit{// BO:}} $\alpha(\mathbf{x}) = \sum_m \sum_n \frac{w^m_t}{N} \, d_{m,n}(\mathbf{x})$ \hfill via \eqref{eq:kendo-bo}--\eqref{eq:hellinger-bo}
    \STATE {\color{gray}\textit{// AL (skip lines 7--9):}} $\alpha(\mathbf{x}) = \sum_m w^m_t \cdot d_m(\mathbf{x})$ \hfill via \eqref{eq:kendo-al}--\eqref{eq:hellinger-al}
    \STATE $\mathbf{x}_{t+1} = \arg\max_{\mathbf{x}} \alpha(\mathbf{x})$
    \STATE Evaluate $f(\mathbf{x}_{t+1})$; \quad $\mathcal{D}_{t+1} = \mathcal{D}_t \cup \{(\mathbf{x}_{t+1}, y_{t+1})\}$
    \FOR{$m \in \mathcal{K}$}
        \STATE Update $w^m_{t+1}$ via \eqref{eq:weight-update}; \quad Update GP posterior mean and variance \hfill via \eqref{eq:mean_m}--\eqref{eq:variance_m} (supplementary)
    \ENDFOR
    \STATE {\color{gray}\textit{// AL only: compute MLL on $\mathcal{V}$}}
\ENDFOR
\RETURN {\color{gray}\textit{BO:}} $\arg\max_{(\mathbf{x},y) \in \mathcal{D}_T} y$ \quad {\color{gray}\textit{AL:}} trained model with $\mathcal{D}_T$
\end{algorithmic}
\end{algorithm}

\subsection{Extensions to Active Learning}
\label{extensions to al}
The optimizer conditioning in KENDO-BO focuses on data collection near promising optima. For active learning, where the goal is global function approximation, we simply remove this conditioning. The KENDO-AL acquisition function measures the disagreement between the ensemble marginal and each component directly
\begin{equation}
\alpha_{\text{SAL}}(x) = \sum_{m=1}^M w^m_t \cdot d(p(y_x \mid \mathcal{D}_t),\, p(y_x \mid i=m, \mathcal{D}_t)) \label{eq:kendo-al}
\end{equation}
where the Hellinger distance between the moment-matched marginal and each kernel's posterior takes the form
\begin{equation}
d_m(x) = 1 - \sqrt{\frac{2\bar{\sigma}_t(x) \sigma_{m,t}(x)}{\bar{\sigma}^2_t(x) + \sigma^2_{m,t}(x)}} \exp\!\left[ -\frac{(\bar{\mu}_t(x) - \mu_{m,t}(x))^2}{4(\bar{\sigma}^2_t(x) + \sigma^2_{m,t}(x))} \right] \label{eq:hellinger-al}
\end{equation}
This formulation prioritizes locations where kernel identity has maximal impact on predictions. Unlike MCMC-based SAL \citep{scorebo}, KENDO-AL avoids sampling overhead while retaining the ability to measure full distributional disagreement. The method handles both outputscale uncertainty (driving global exploration) and lengthscale uncertainty (encouraging repeated queries for noise estimation), as each kernel's contribution is weighted by its predictive likelihood. The unified process is shown in Algorithm~\ref{alg:egp-unified}.

\section{KENDO for MOBO}
\label{KENDO for MOBO}
In this section, we extend the KENDO framework to the multi-objective Bayesian optimization (MOBO) setting, where $K$ (possibly conflicting) objectives $\{f^k\}_{k=1}^K$ are optimized simultaneously. Unlike the single-objective case, the solution is no longer a single optimizer but the Pareto set $\mathcal{X}^* = \{\mathbf{x}^* \in \mathcal{X} : \nexists\, \mathbf{x} \in \mathcal{X}\ \text{s.t.}\ \mathbf{f}({\bf x}) \succ \mathbf{f}({\bf x}^*)\}$, $\mathbf{f}(\mathbf{x}):= [f^1(\mathbf{x}),\ldots,f^K(\mathbf{x})]^\top$ ,  formed under the Pareto dominance relation $\succ$, where $\mathbf{f}({\bf x}) \succ \mathbf{f}({\bf x}^*)$ denotes that $f^k({\bf x}) \geq f^k({\bf x}^*)$ for all $k$ with strict inequality for at least one $k$ in maximization problem. This relation induces the Pareto front $\mathcal{P}^* = \{\mathbf{f}({\bf x}^*) : {\bf x}^* \in \mathcal{X}^*\}$ in objective space, and the discovered front is commonly evaluated against $\mathcal{P}^*$ via the hypervolume indicator with respect to a reference point~\citep{emmerich2006single}. Therefore, two design choices of KENDO-BO are key in the MOBO setting. The kernel ensemble needs to be instantiated per objective, allowing different $f^k$ to prefer different kernels in $\mathcal{K}$, and the optimizer conditioning that drives the disagreement signal in \eqref{eq:kendo-bo} must be extended to the Pareto setting, where the solution is inherently a set rather than a single point.
For $K$ objectives $\{f^k\}_{k=1}^K$, we maintain independent EGPs per objective with weights $w_t^{k,m}$. {Extending to multi-objective optimization, however, requires careful treatment, as the notion of a single ``optimizer'' no longer applies; the solution is a Pareto front rather than a single point.} A direct extension would condition on sampled Pareto fronts, but this introduces $P_{m,n}$ phantom observations simultaneously, diffusing the conditioning signal across the input space and fundamentally departing from the single-optimizer mechanism that makes SCoreBO effective. Our experiments confirm that this yields suboptimal performance (Section \ref{MOBOresults}).

Instead, we propose a random scalarization strategy that preserves the single-optimizer structure. At each iteration, we sample $S$ weight vectors $\boldsymbol{\lambda}_s \sim \text{Dir}(\mathbf{1}_K)$ from a symmetric Dirichlet distribution. For each kernel $m$ and scalarization $s$, we draw function paths $f^k_{m,s}$ for all objectives via Random Fourier features (RFF) \citep{rahimi2007random, wilson2020efficiently}, form the scalarized objective $g_{m,s}({\bf x}) = \sum_k \lambda_{s,k} f^k_{m,s}({\bf x})$, and find its single optimizer ${\bf x}^*_{m,s}$. We then condition each per-objective GP on the phantom observation $({\bf x}^*_{m,s}, f^{*,k}_{m,s})$ where $f^{*,k}_{m,s} = f^k_{m,s}({\bf x}^*_{m,s})$, and aggregate per-objective Hellinger distances for the acquisition step as follows:
\begin{equation}
    \alpha({\bf x}) = \sum_{k=1}^{K} \sum_{m=1}^{M} w_t^{k,m} \cdot \frac{1}{S} \sum_{s=1}^{S} d\!\left(\bar{p}(y_x^k \mid \mathcal{D}_t),\; p(y_x^k \mid i\!=\!m,\, (\mathbf{x}^*_{m,s}, f^{*,k}_{m,s}),\, \mathcal{D}_t)\right) \label{eq:kendo-mo}
\end{equation}
where $\bar{p}(y_x^k \mid \mathcal{D}_t)$ is the moment-matched marginal for objective $k$ and $d(\cdot,\cdot)$ is the Hellinger distance \eqref{eq:hellinger-bo}.

This design preserves the focused, single-point conditioning that makes KENDO-BO effective, while a sufficient diversity of scalarization directions $\{\boldsymbol{\lambda}_s\}$ provides coverage of the full Pareto front; Appendix~\ref{app:ablation-dirichlet} confirms that the framework is robust to the choice of Dirichlet concentration controlling this diversity. Finding each scalarized optimizer requires only single-objective optimization via L-BFGS on the RFF path, avoiding multi-objective solvers such as NSGA-II~\citep{deb2002nsga2}. The full process is summarized in Algorithm~\ref{alg:KENDO-MO} in the supplementary file.

\section{Experiments}
\label{sec:experiments}

\subsection{Experimental Setup}
\label{sec:exp-setup}
All EGP-based methods use $\mathcal{K} = \{\text{RBF}, \text{Mat\'{e}rn-1.5}, \text{Mat\'{e}rn-2.5}\}$ with per-kernel marginal-likelihood fitting. To ensure fair comparison, every method (including baselines) uses the same GP priors: $\text{GammaPrior}(3.0, 6.0)$ on lengthscale and $\text{GammaPrior}(2.0, 0.15)$ on outputscale. For single-objective BO, we compare against NEI~\citep{letham2019nei}, PES~\citep{hernandez2014predictive}, GIBBON (MES)~\citep{moss2021gibbon}, JES~\citep{hvarfner2022joint}, the MCMC-based SCoreBO~\citep{scorebo}, and EGP-TS~\citep{lu2023surrogate} that uses the same kernel ensemble as ours but replaces the disagreement-based acquisition with standard Thompson sampling. For multi-objective BO, baselines include JESMO~\citep{tu2022jesmo}, MESMO~\citep{belakaria2019mesmo}, PES(MO), qNEHVI~\citep{daulton2021qnehvi}, EGP-TS-MOBO, and KENDO-MO-PF (our Pareto-front-conditioning variant from Section \ref{KENDO for MOBO}, included to validate the scalarization design). For active learning, we compare against BALD~\citep{houlsby2011bayesian}, BQBC~\citep{riis2022bayesian}, QBMGP~\citep{riis2022bayesian}, and SAL~\citep{scorebo}; all SAL variants use Hellinger distance.

The benchmarks used for evaluation are categorized as follows. \emph{Single-objective BO:} Branin (2D), Rosenbrock (2D, 4D), Hartmann (3D, 4D, 6D), and three additional real world problems, RobotPush (3D), RF-HPO on Adult (4D), and Borehole (8D), to cover synthetic, simulator, and HPO problems. \emph{Multi-objective BO:} two synthetic functions: ZDT2~\citep{zitzler2000zdt} ($d{=}6$, $K{=}2$), DTLZ2~\citep{deb2002dtlz} ($d{=}6$, $K{=}3$), and four real world problems: VehicleSafety ($d{=}5$, $K{=}3$), CarSideImpact ($d{=}7$, $K{=}3$), Penicillin~\citep{liang2021penicillin} ($d{=}7$, $K{=}3$), and LCBench~\citep{zimmer2021autopytorch} ($d{=}7$, $K{=}2$), reported via $\log_{10}$ hypervolume difference. \emph{Active learning:} Higdon (1D), Branin (2D), Ishigami (3D) and Hartmann (6D), along with three real benchmarks, RobotPush (3D), GBT-HPO (4D) and Airfoil (5D), reported via negative MLL on held-out validation sets, with relative ranking as an aggregate view. All curves are averaged over 25 seeds; results for single-kernel baselines are additionally averaged over the three kernel variants in $\mathcal{K}$.

\subsection{Single-Objective Optimization Results}
\label{sec:exp-so}
\begin{figure}[t]
    \centering
    \includegraphics[width=\textwidth]{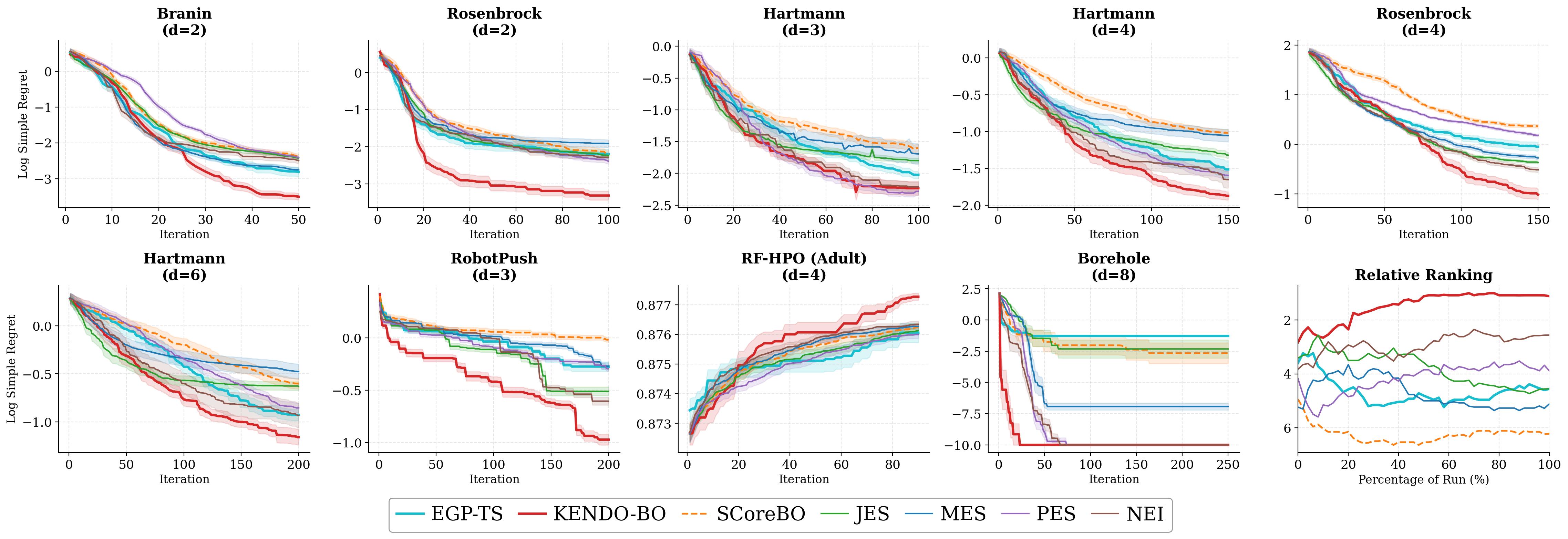}
    \caption{Average $\log_{10}$ simple regret on nine single-objective benchmarks plus the relative ranking panel, averaged over 25 random seeds. Results are averaged over 3 kernel variants. On Borehole, curves saturate at $-10$ due to a hard floor on $\log_{10}$ regret; KENDO-BO reaches the floor earliest.}
    \label{fig:single_objective}
\end{figure}

For performance evaluation, the central question is whether substituting MCMC with a discrete kernel ensemble degrades optimization quality. Figure~\ref{fig:single_objective} shows that this well-motivated replacement does not degrade optimization performance. On the contrary, across all nine benchmarks, spanning synthetic functions to real problems and low to high dimensional regimes, KENDO-BO matches or outperforms all baselines. The relative-ranking panel confirms that it maintains the top average rank as the budget grows.
The comparison against EGP-TS isolates the contribution of the acquisition function, since both methods share the same kernel ensemble and weighting scheme. KENDO-BO consistently outperforms EGP-TS, indicating that the gain comes from disagreement-based, optimizer-conditioned querying rather than from the ensemble alone.

\subsection{Multi-Objective Optimization Results}
\label{MOBOresults}
\begin{figure}[t]
    \centering
    \includegraphics[width=\textwidth]{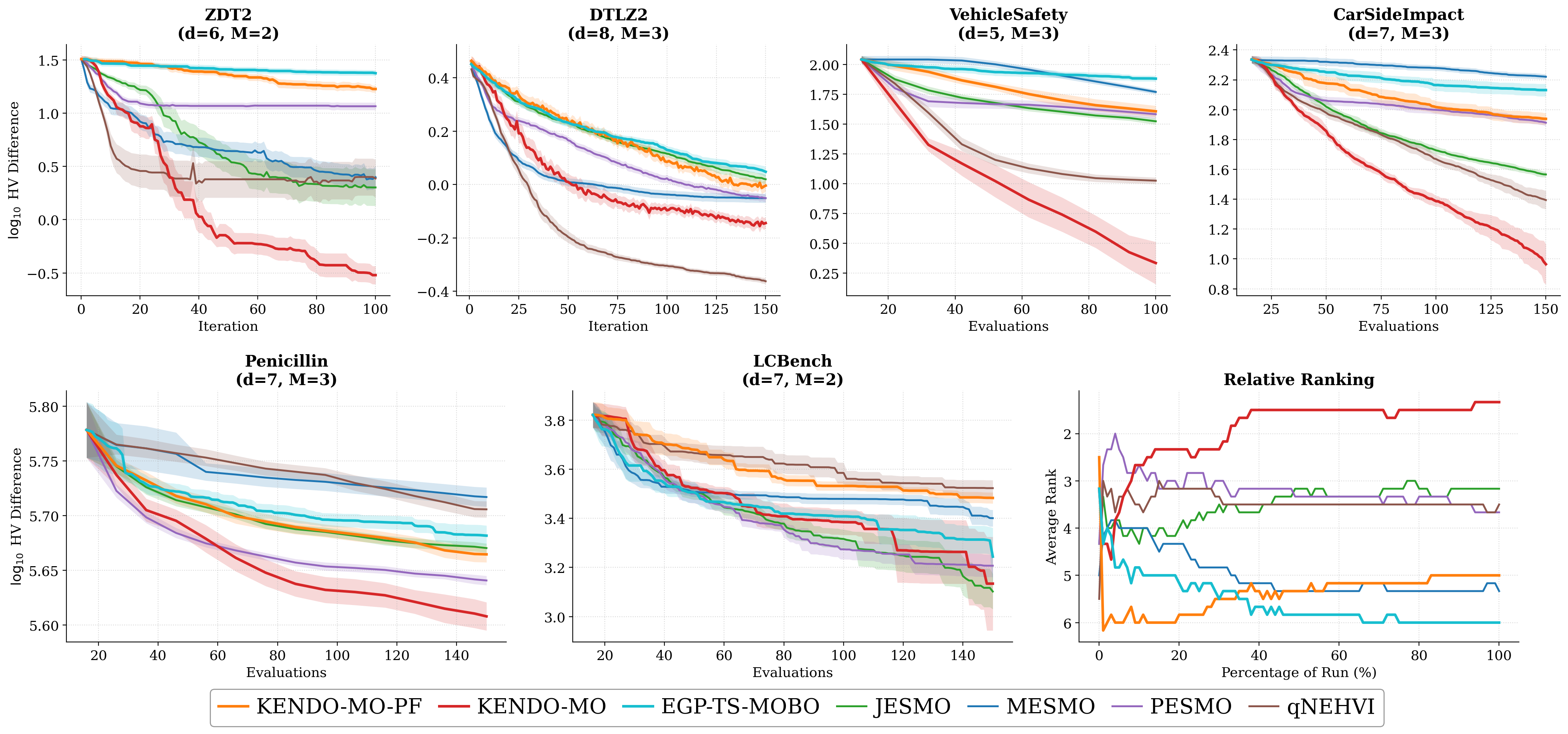}
    \caption{Multi-objective optimization results on six benchmarks plus the relative ranking panel. Reported as $\log_{10}$ hypervolume difference, averaged over 25 random seeds and 3 kernel variants.}
    \label{fig:multi_objective}
\end{figure}

Figure~\ref{fig:multi_objective} reports results on six MOBO benchmarks. KENDO-MO achieves the lowest $\log_{10}$ hypervolume difference on the majority of problems, with the largest margins on Penicillin and CarSideImpact, both real-world problems whose objectives have heterogeneous landscapes that benefit from per-objective adaptive weighting. The aggregate ranking panel further shows KENDO-MO on top across the full budget.

In comparison with KENDO-MO-PF, which conditions on sampled Pareto fronts instead of scalarized optimizers, KENDO-MO-PF underperforms KENDO-MO consistently across all six benchmarks. This corroborates what we discuss in Section \ref{KENDO for MOBO}: conditioning on an entire Pareto set introduces many phantom observations at once, and the conditioning signal gets diffused across the input space rather than concentrated where it is informative. Random scalarization keeps the focused, single-point conditioning that makes single-objective KENDO-BO work, while the diversity of $\boldsymbol{\lambda}_s$ ensures the Pareto front is still adequately covered.

\subsection{Active Learning Results}
\label{sec:exp-al}
\begin{figure}[t]
    \centering
    \includegraphics[width=\textwidth]{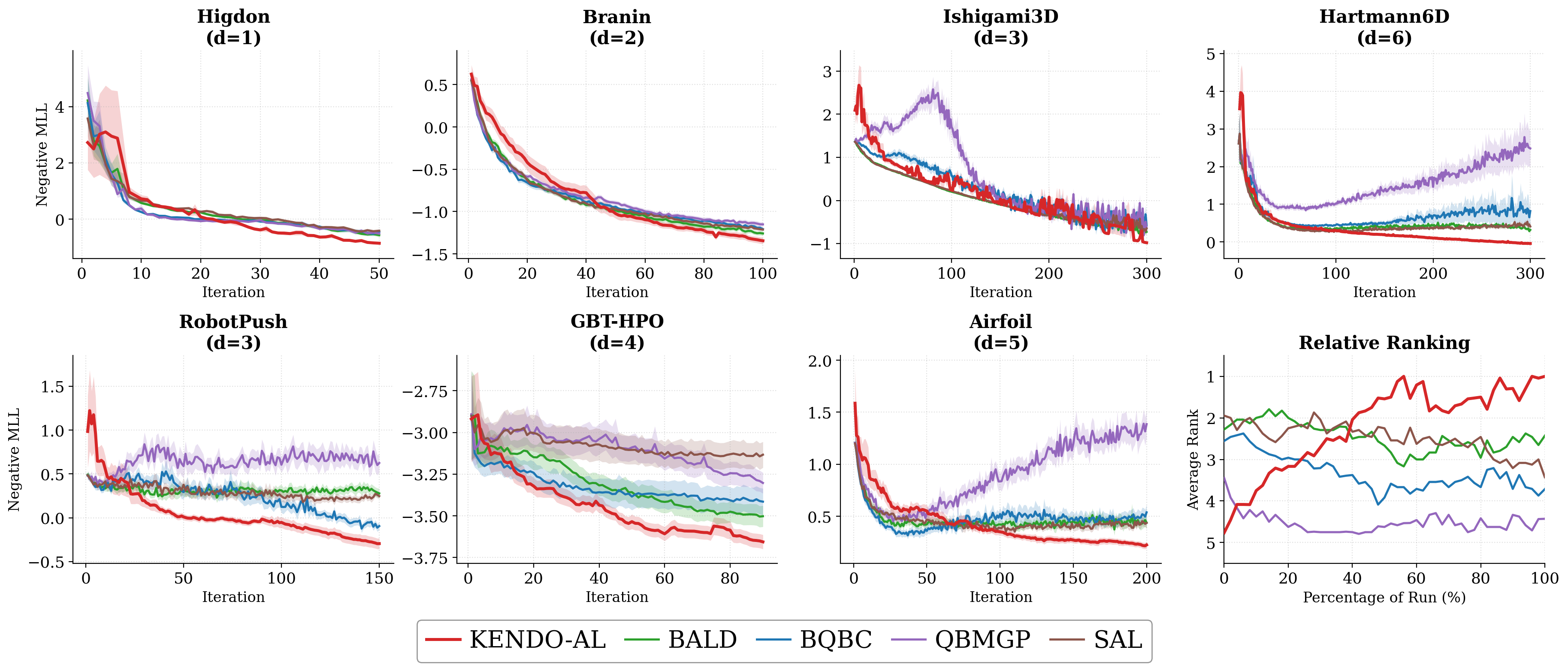}
    \caption{Active learning results on seven benchmarks plus the relative ranking panel. Negative MLL on held-out validation sets, averaged over 25 random seeds and three kernel variants.}
    \label{fig:active_learning_results}
\end{figure}

Unlike BO, active learning calls for pure exploration, where the goal is global predictive accuracy, not finding an optimum. This objective allows us to test whether the ensemble's kernel diversity provides intrinsic value beyond its role in guiding optimization. Figure~\ref{fig:active_learning_results} evaluates KENDO-AL on seven benchmarks spanning synthetic functions (Higdon, Branin, Ishigami, Hartmann), a simulator task (RobotPush), an HPO problem (GBT-HPO), and an engineering surrogate (Airfoil).
KENDO-AL achieves the lowest negative MLL on every benchmark and holds the best average rank in the aggregate panel. The disagreement-based methods (KENDO-AL and SAL) consistently perform well and KENDO-AL further improves on SAL by replacing MCMC over hyperparameters with the kernel ensemble, which bypasses the sensitivity to a pre-selected kernel that QBMGP and BQBC suffer from. These gains come with a $7.6\times$--$26.9\times$ speedup over SAL (Appendix~\ref{app:ablation-efficiency}).

\subsection{Ablation Studies}
\label{ablation studies}
We evaluate four design choices of KENDO: adaptive versus uniform weighting, computational efficiency, kernel weight dynamics, and the choice of scalarization for MOBO. Detailed results and discussion are deferred to Appendix~\ref{app:ablation}; we summarize the main findings here.

\textbf{Adaptive vs.\ uniform weighting.} The Bayesian weighting scheme in \eqref{eq:weight-update} outperforms uniform averaging across all three task families, with the gap pronounced on benchmarks where the preferred kernel varies across input regions (Appendix~\ref{app:ablation-weighting}).

\textbf{Computational efficiency.} KENDO runs $3.1\times$--$5.4\times$ faster than SCoreBO, and KENDO-AL runs $7.6\times$--$26.9\times$ faster than SAL; the weight update itself contributes less than $7$\,ms per iteration (Appendix~\ref{app:ablation-efficiency}).

\textbf{Kernel weight convergence.} The ensemble exhibits winner-take-all behavior on benchmarks with a clearly superior kernel, and remains diversified when no single kernel dominates (Appendix~\ref{app:ablation-weights}).

\textbf{Sensitivity to scalarization design.} The MOBO scalarization layer is
robust to both the scalarization form and the Dirichlet concentration
(Appendices~\ref{app:ablation-scalarization}--\ref{app:ablation-dirichlet}).

\section{Conclusions}
\label{sec:conclusion}

The central focus of this paper is kernel family diversity as a source of model uncertainty. {We address this by replacing hyperparameter sampling from costly fully Bayesian MCMC based approaches with an explicit kernel ensemble and adaptive Bayesian weighting, that can be readily leveraged for both Bayesian optimization and active learning.
Our KENDO-BO and KENDO-AL variants achieve superior optimization and predictive performance over MCMC-based approaches with substantial computational savings.}
The proposed approach naturally extends to multi-objective optimization via random scalarization, while ablation studies corroborate the effectiveness of adaptive weighting and the automatic model selection capabilities of the ensemble approach.

\textbf{Limitations.} The current framework relies on moment matching to approximate Gaussian mixture posteriors, which may underestimate uncertainty in cases with highly disparate kernel predictions. Additionally, the kernel dictionary requires manual specification; though automatic construction via kernel composition \citep{duvenaud2013structure} could be explored.

\textbf{Future Work.} Our future research agenda includes: (1) adaptive kernel dictionary construction during optimization; (2) integration with batch BO for parallel evaluations \citep{gonzalez2016batch}; (3) extension to constrained optimization with feasibility modeling per kernel; (4) application to hyperparameter tuning of deep neural networks; and (5) investigation of the ensemble mechanism in deep active learning contexts where model capacity exceeds kernel expressiveness.


\bibliographystyle{abbrvnat}
\bibliography{references}

\newpage
\appendix

\begin{center}
    \textbf{\Large Supplementary File}
\end{center}

\section{KENDO-MO Algorithm for Multi-Objective Optimization}
\label{app:kendo-mo-alg}

This appendix presents the full procedure of KENDO-MO described in Section \ref{KENDO for MOBO}.

\begin{algorithm}[H]
\caption{KENDO-MO for Multi-Objective Optimization}
\label{alg:KENDO-MO}
\small
\begin{algorithmic}[1]
\REQUIRE Kernel dict.\ $\mathcal{K}$, \#scalarizations $S$, \#objectives $K$, data $\mathcal{D}_{0}$, budget $T$
\FOR{$k = 1, \ldots, K$; $m \in \mathcal{K}$}
    \STATE Fit $\alpha_m^k$ via $\max p(\mathcal{D}_{0}^k | i=m, \alpha)$; \quad Init $w_0^{k,m} = 1/M$
\ENDFOR
\FOR{$t = 0, \ldots, T-1$}
    \FOR{$k = 1, \ldots, K$}
        \STATE Compute $\bar{\mu}_t^k(\cdot), \bar{\sigma}^{2,k}_t(\cdot)$ via \eqref{eq:mean}--\eqref{eq:var} with weights $w_t^{k,m}$
    \ENDFOR
    \FOR{$m \in \mathcal{K}$; $s = 1, \ldots, S$}
        \STATE Sample $\boldsymbol{\lambda}_s \sim \text{Dir}(\mathbf{1}_K)$; \quad Sample $f^k_{m,s} \sim p(f^k | i=m, \mathcal{D}_t)$ via RFF for all $k$
        \STATE $g_{m,s}(x) = \sum_k \lambda_{s,k} f^k_{m,s}(\mathbf{x})$; \quad Find $x^*_{m,s} = \arg\max_x g_{m,s}(\mathbf{x})$
        \FOR{$k = 1, \ldots, K$}
            \STATE Condition GP $m$ on $(\mathbf{x}^*_{m,s}, f^k_{m,s}(\mathbf{x}^*_{m,s}))$ to get $\mu^{*,k}_{m,s}, \sigma^{2,*,k}_{m,s}$
        \ENDFOR
    \ENDFOR
    \STATE $\mathbf{x}_{t+1} = \arg\max_{x} \sum_{k} \sum_{m} w_t^{k,m} \cdot \frac{1}{S} \sum_{s} d(\bar{p}^k, p^k_{m,s})$ \hfill via \eqref{eq:kendo-mo}
    \STATE Evaluate $\mathbf{y}_{t+1} = (f^1(\mathbf{x}_{t+1}), \ldots, f^K(\mathbf{x}_{t+1}))$; \quad $\mathcal{D}_{t+1} = \mathcal{D}_t \cup \{(\mathbf{x}_{t+1}, \mathbf{y}_{t+1})\}$
    \FOR{$k = 1, \ldots, K$; $m \in \mathcal{K}$}
        \STATE Update $w_{t+1}^{k,m}$ via \eqref{eq:weight-update}; \quad Update GP posterior \hfill via \eqref{eq:mean_m}--\eqref{eq:variance_m}
    \ENDFOR
\ENDFOR
\RETURN Pareto front from $\mathcal{D}_T$
\end{algorithmic}
\end{algorithm}

\section{Derivation of the KENDO-BO Acquisition Function}
\label{app:derivation}

This appendix details how the KENDO-BO acquisition function in \eqref{eq:kendo-bo} arises from the disagreement principle of SCoreBO~\citep{scorebo} once the source of model uncertainty is changed from continuous hyperparameters to a discrete kernel posterior. We also cover the moment-matching step that produces the Gaussian summary in \eqref{eq:mean} and \eqref{eq:var}, and the closed-form Hellinger distance in \eqref{eq:hellinger-bo}.

\subsection{From SCoreBO's continuous expectation to KENDO's discrete sum}
\label{app:deriv-replacement}

SCoreBO defines its acquisition as the expected Hellinger disagreement between the marginal predictive and an optimizer-conditioned predictive
\begin{equation}
\alpha_{\text{SC}}(\mathbf{x}) = \mathbb{E}_{\boldsymbol{\theta}, \ast}\!\left[d\!\left(p(y_\mathbf{x}\mid \mathcal{D}_t),\; p(y_\mathbf{x}\mid \boldsymbol{\theta}, \ast, \mathcal{D}_t)\right)\right],
\label{eq:scorebo-recap}
\end{equation}
where the joint posterior factors as $p(\boldsymbol{\theta}, \ast \mid \mathcal{D}_t) = p(\ast \mid \boldsymbol{\theta}, \mathcal{D}_t)\, p(\boldsymbol{\theta} \mid \mathcal{D}_t)$. The outer expectation is intractable, and SCoreBO approximates it by drawing hyperparameter samples $\boldsymbol{\theta}_m$ from $p(\boldsymbol{\theta}\mid \mathcal{D}_t)$ via MCMC, then drawing $N$ optimizers per sample, and applying the uniformly weighted estimator $\frac{1}{MN}\sum_m \sum_n d(\cdot, \cdot)$.

KENDO changes the random quantity itself. Instead of a continuous distribution over hyperparameters of a single kernel, the source of model uncertainty is the kernel \emph{family}, whose posterior is the discrete distribution $p(m \mid \mathcal{D}_t) = w^m_t$ defined by the Bayesian model average in \eqref{eq:weight-update}. The acquisition function then takes the form
\begin{equation}
\alpha(\mathbf{x}) = \mathbb{E}_{m, \ast}\!\left[d\!\left(p(y_\mathbf{x}\mid \mathcal{D}_t),\; p(y_\mathbf{x}\mid m, \ast, \mathcal{D}_t)\right)\right].
\end{equation}
Factoring the joint posterior as $p(m, \ast \mid \mathcal{D}_t) = p(\ast \mid m, \mathcal{D}_t)\, w^m_t$ and expanding the now discrete outer expectation gives
\begin{align}
\alpha(\mathbf{x})
&= \sum_{m=1}^M w^m_t \cdot \mathbb{E}_{\ast \mid m}\!\left[d\!\left(p(y_\mathbf{x}\mid \mathcal{D}_t),\; p(y_\mathbf{x}\mid m, \ast, \mathcal{D}_t)\right)\right] \label{eq:deriv-discrete}\\
&\approx \sum_{m=1}^M w^m_t \cdot \frac{1}{N}\sum_{n=1}^N d_{m,n}(\mathbf{x})
\;=\; \sum_{m=1}^M \sum_{n=1}^N \frac{w^m_t}{N}\, d_{m,n}(\mathbf{x}),
\label{eq:deriv-mc}
\end{align}
which recovers \eqref{eq:kendo-bo}. The inner step uses $N$ optimizer samples $\ast_{m,n} = (\mathbf{x}^*_{m,n}, f^*_{m,n})$ obtained by optimizing function paths $f_{m,n} \sim p(f \mid m, \mathcal{D}_t)$ generated through random Fourier features~\citep{rahimi2007random, wilson2020efficiently}. Conditioning the $m$-th GP on the phantom observation $(\mathbf{x}^*_{m,n}, f^*_{m,n})$ yields the conditional moments $\mu^{\ast_{m,n}}_{m,t}$ and $\sigma^{2,\ast_{m,n}}_{m,t}$.

\subsection{Moment matching of the marginal mixture}
\label{app:deriv-mm}

The marginal predictive in \eqref{eq:gp-mixture} is a Gaussian mixture, so its Hellinger distance to a single Gaussian conditional has no closed form expression. We replace the mixture by a single Gaussian whose first and second moments match those of the mixture exactly. For the mixture $\sum_m w^m_t \mathcal{N}(\mu_{m,t}, \sigma^2_{m,t})$, these are
\begin{align}
\bar{\mu}_t(\mathbf{x}) &= \sum_{m=1}^M w^m_t \mu_{m,t}(\mathbf{x}), \\
\bar{\sigma}^2_t(\mathbf{x}) &= \sum_{m=1}^M w^m_t \sigma^2_{m,t}(\mathbf{x}) + \sum_{m=1}^M w^m_t \big(\mu_{m,t}(\mathbf{x}) - \bar{\mu}_t(\mathbf{x})\big)^2,
\end{align}
which match \eqref{eq:mean} and \eqref{eq:var}. The variance decomposes additively into a within-component term (average individual variance) and a between-component term (spread of component means around the mixture mean).

\subsection{Closed-form Hellinger distance between two Gaussians}
\label{app:deriv-hellinger}

For two univariate Gaussians $z_1 \sim \mathcal{N}(\mu_1, \sigma_1^2)$ and $z_2 \sim \mathcal{N}(\mu_2, \sigma_2^2)$, the squared Hellinger distance admits the closed form
\begin{equation}
H^2(z_1, z_2) = 1 - \sqrt{\frac{2\sigma_1 \sigma_2}{\sigma_1^2 + \sigma_2^2}} \exp\!\left[-\frac{1}{4}\frac{(\mu_1 - \mu_2)^2}{\sigma_1^2 + \sigma_2^2}\right].
\label{eq:hellinger-gauss}
\end{equation}
Setting $z_1 \sim \mathcal{N}(\bar{\mu}_t(\mathbf{x}), \bar{\sigma}^2_t(\mathbf{x}))$ as the moment-matched marginal and $z_2 \sim \mathcal{N}(\mu^{\ast_{m,n}}_{m,t}(\mathbf{x}), \sigma^{2,\ast_{m,n}}_{m,t}(\mathbf{x}))$ as the conditional posterior of the $m$-th GP given optimizer sample $n$, and writing $d_{m,n}(\mathbf{x}) = H(z_1, z_2)$, we recover the per-pair Hellinger distance in \eqref{eq:hellinger-bo}.

\subsection{Reduction to KENDO-AL}
\label{app:deriv-al}

Removing the optimizer conditioning leaves
\begin{equation}
\alpha_{\text{AL}}(\mathbf{x}) = \mathbb{E}_m\!\left[d\!\left(p(y_\mathbf{x}\mid \mathcal{D}_t),\; p(y_\mathbf{x}\mid m, \mathcal{D}_t)\right)\right] = \sum_{m=1}^M w^m_t \, d_m(\mathbf{x}),
\end{equation}
where $d_m(\mathbf{x})$ uses \eqref{eq:hellinger-gauss} between the moment-matched marginal and the $m$-th component. This recovers \eqref{eq:kendo-al}, and \eqref{eq:hellinger-al} follows from the same closed form.

\section{Ablation Studies}
\label{app:ablation}

This appendix presents the full ablation studies summarized in Section \ref{ablation studies} of the main text.

\subsection{Adaptive vs.\ Uniform Weighting}
\label{app:ablation-weighting}

\begin{figure}[!t]
    \centering
    \includegraphics[height=0.38\textheight]{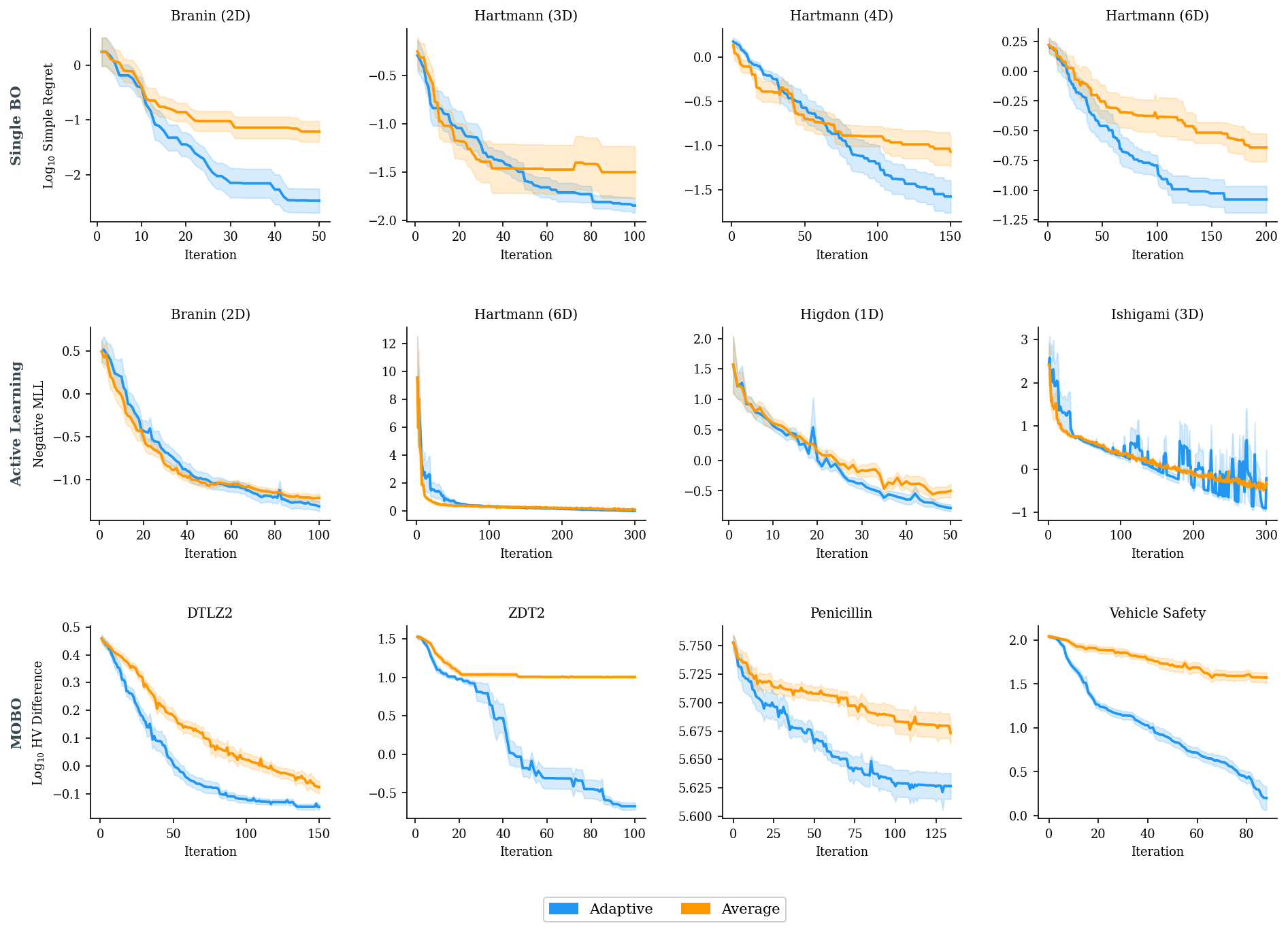}
    \caption{Convergence comparison between adaptive Bayesian weighting and uniform (average) weighting across all three task settings. \textbf{Top row:} $\log_{10}$ simple regret on single-objective benchmarks. \textbf{Middle row:} Negative MLL on active learning benchmarks. \textbf{Bottom row:} $\log_{10}$ hypervolume difference on multi-objective benchmarks. Adaptive weighting consistently achieves better performance.}
    \label{fig:convergence_ablation}
\end{figure}

Figure~\ref{fig:convergence_ablation} compares KENDO with adaptive Bayesian weights \eqref{eq:weight-update} against a variant using fixed uniform weights ($w^m = 1/M$ throughout optimization). Across all three task categories, single-objective BO, active learning, and multi-objective optimization, adaptive weighting consistently achieves lower final regret.

On single-objective benchmarks, the advantage is most pronounced on Branin (2D) and Hartmann (6D), where adaptive weighting achieves roughly $0.5$ and $0.3$ lower $\log_{10}$ regret at convergence, respectively. For active learning, adaptive weighting leads to substantially better MLL on Hartmann (6D) and Ishigami (3D), where kernel preference varies significantly across regions of the input space. In multi-objective settings, the advantage is evident on DTLZ2 and Penicillin, where different objectives may favor different kernels, and adaptive per-objective weighting captures this structure more effectively than uniform averaging. These results confirm that the Bayesian weight update mechanism provides measurable performance gains by dynamically allocating model capacity to the most predictive kernels.

\subsection{Computational Efficiency}
\label{app:ablation-efficiency}

\begin{figure}[t]
    \centering
    \includegraphics[width=\textwidth]{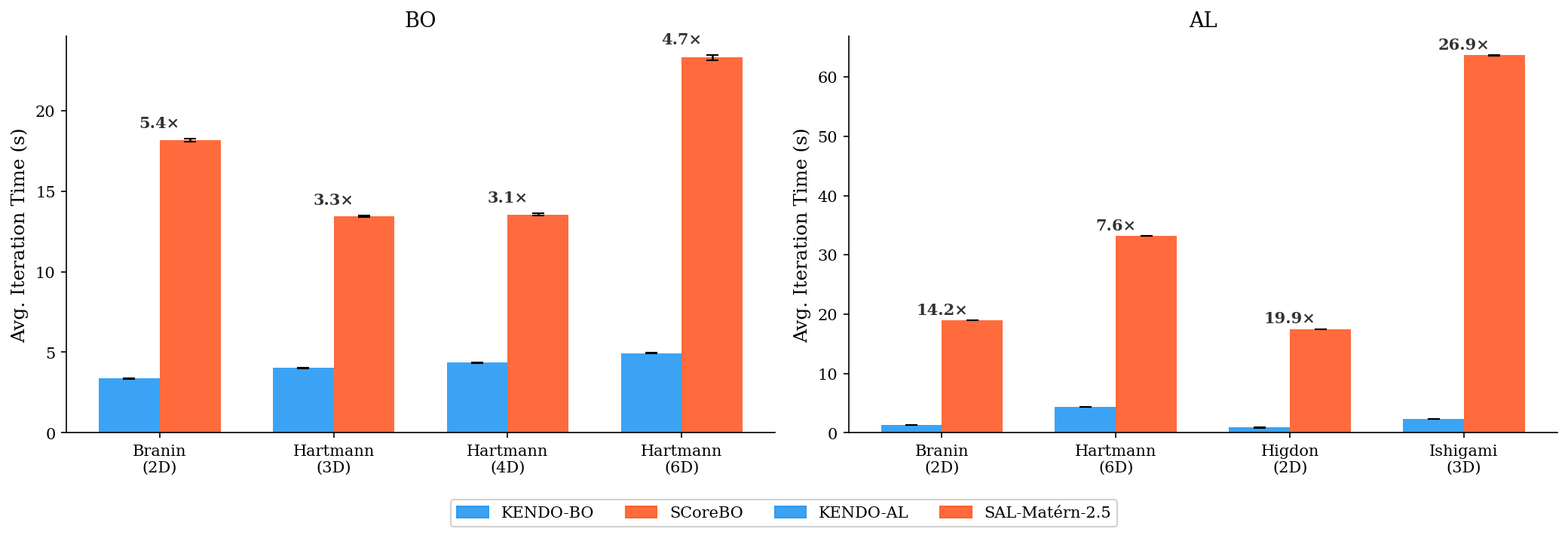}
    \caption{Per-iteration wall-clock time comparison. \textbf{Left:} KENDO vs.\ SCoreBO on single-objective benchmarks ($3.1\times$--$5.4\times$ speedup). \textbf{Right:} KENDO-AL vs.\ SAL-Mat\'{e}rn-2.5 on active learning benchmarks ($7.6\times$--$26.9\times$ speedup). The ensemble mechanism eliminates MCMC overhead while maintaining competitive optimization and learning performance (cf.\ Figures~\ref{fig:single_objective}--\ref{fig:active_learning_results}).}
    \label{fig:time_comparison}
\end{figure}

A central claim of KENDO is that replacing MCMC hyperparameter sampling with an explicit kernel ensemble yields substantial computational savings. Figure~\ref{fig:time_comparison} validates this claim by comparing per-iteration wall-clock time between EGP methods and their MCMC-based counterparts.

For Bayesian optimization, KENDO achieves $3.1\times$--$5.4\times$ speedup over SCoreBO across benchmarks, with the largest gain on Branin (2D) where MCMC overhead dominates relative to the simple acquisition landscape. The speedup is consistent across problem dimensions, ranging from $3.1\times$ on Hartmann (4D) to $5.4\times$ on Branin. The computational advantage is even more dramatic for active learning. KENDO-AL achieves $7.6\times$--$26.9\times$ speedup over SAL-Mat\'{e}rn-2.5, with the largest gains on Ishigami ($26.9\times$) and Higdon ($19.9\times$). This amplified speedup arises because active learning acquisition functions (which do not require RFF-based function sampling and optimizer search) have lower base cost, making the relative contribution of MCMC overhead proportionally larger.

\begin{figure}[t]
    \centering
    \includegraphics[width=\textwidth]{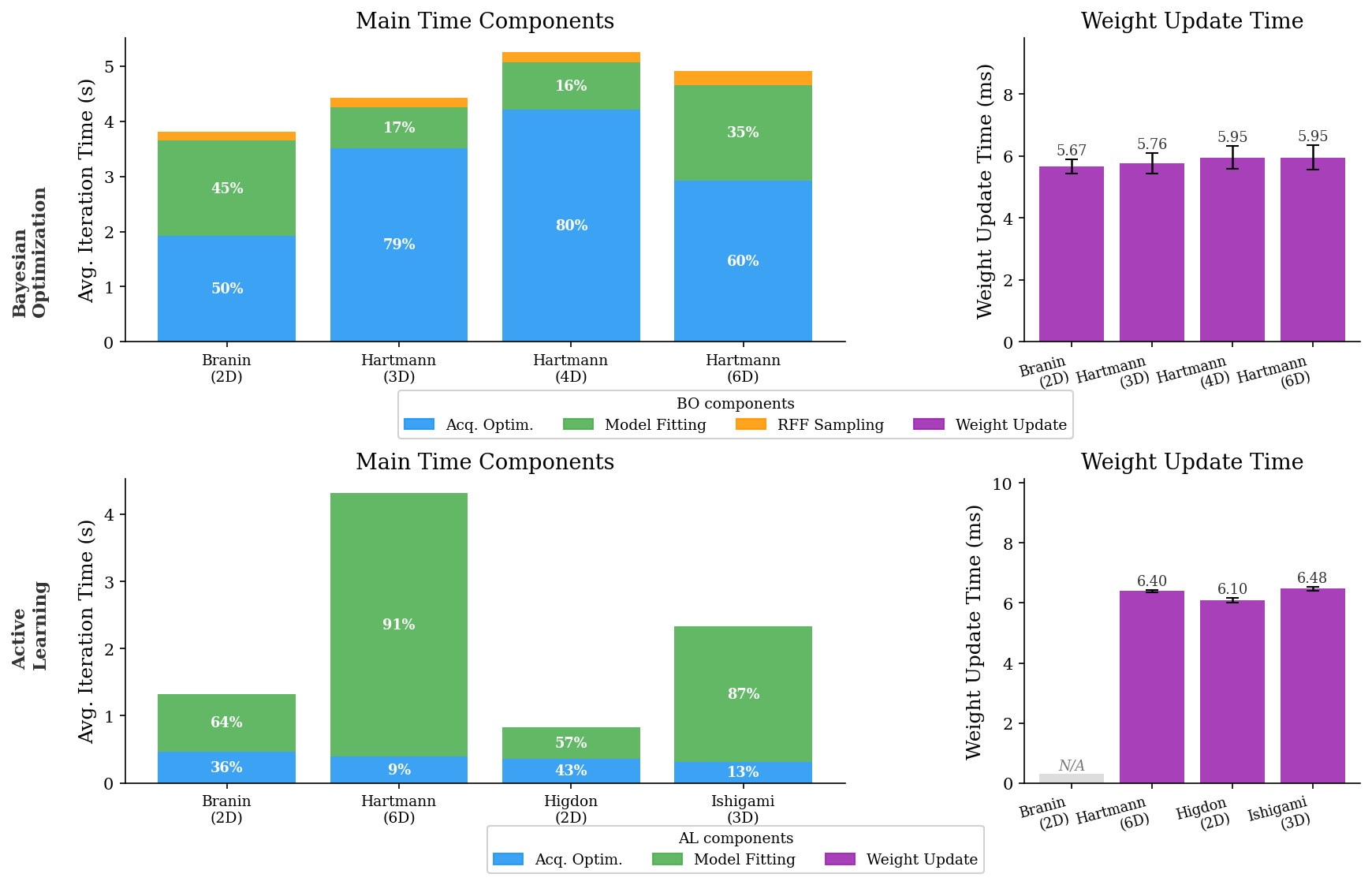}
    \caption{Time component breakdown for KENDO (\textbf{top}) and KENDO-AL (\textbf{bottom}). Left panels show main time components (acquisition optimization, model fitting, RFF sampling, weight update). Right panels isolate weight update time, confirming it contributes negligibly ($<7$\,ms per iteration). For BO, acquisition optimization dominates (50--80\%); for AL, model fitting becomes the primary cost on higher-dimensional benchmarks.}
    \label{fig:time_breakdown}
\end{figure}

Figure~\ref{fig:time_breakdown} provides a detailed breakdown of where computation time is spent within KENDO and KENDO-AL. For BO, acquisition function optimization dominates (50--80\% of iteration time), with model fitting consuming most of the remainder (17--45\%). RFF sampling and weight updates contribute negligibly. For active learning, the split between acquisition optimization and model fitting varies by benchmark, but weight update time remains consistently below 7 milliseconds, confirming that the Bayesian weight update \eqref{eq:weight-update} introduces negligible overhead.

\subsection{Kernel Weight Convergence and Winner-Take-All Behavior}
\label{app:ablation-weights}

\begin{figure}[t]
    \centering
    \includegraphics[width=\textwidth]{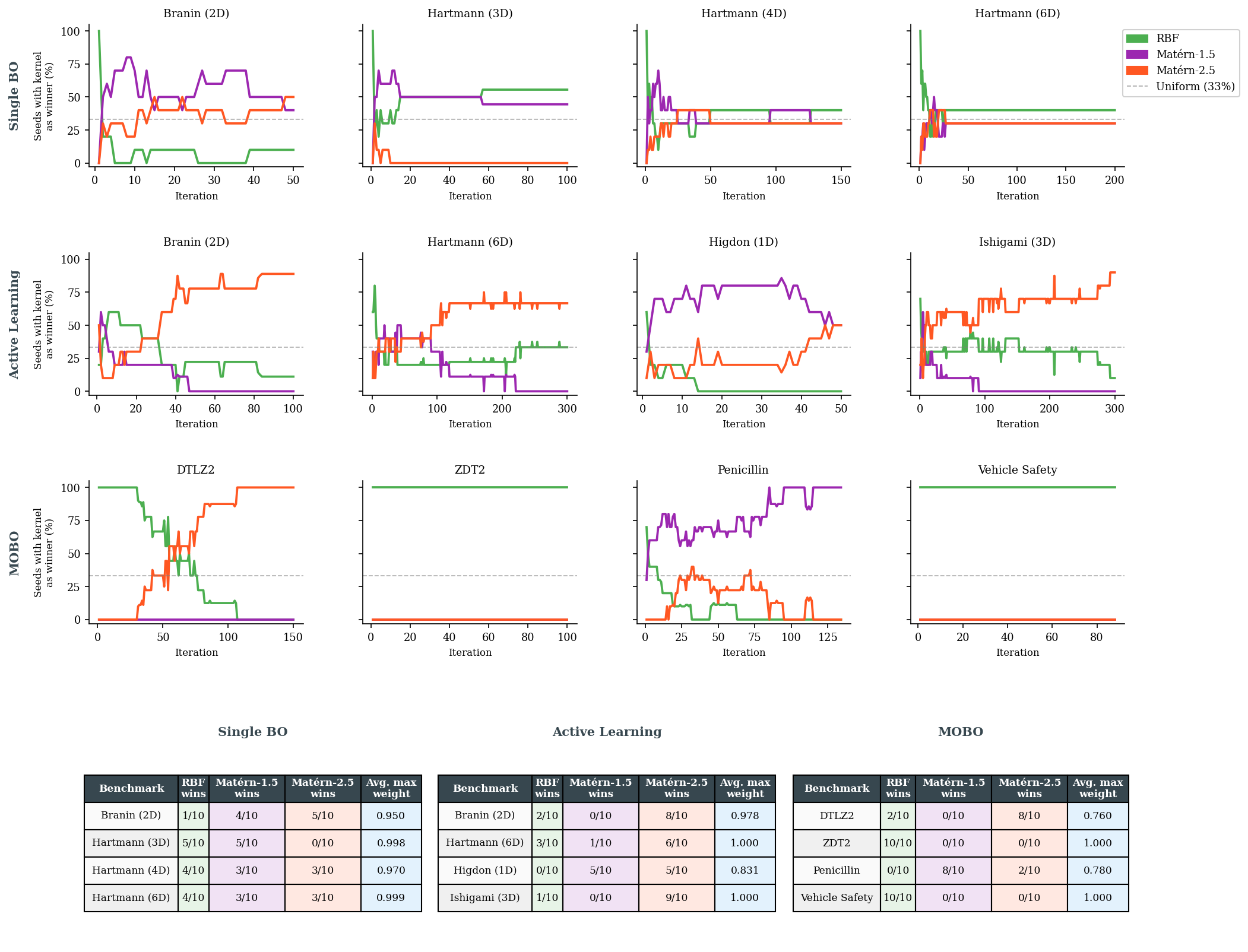}
    \caption{Kernel weight dynamics across all benchmarks. Line plots show the fraction of seeds (out of 10) for which each kernel holds the largest weight at each iteration. The dashed gray line indicates the uniform baseline (33\%). Tables summarize the final kernel preference distribution and average maximum weight. The ensemble exhibits decisive model selection on benchmarks with a clear best kernel (e.g., RBF on ZDT2 with max weight 1.000), while maintaining diversity under ambiguity (e.g., Hartmann 3D in BO with a 5/5 split between RBF and Mat\'{e}rn-1.5).}
    \label{fig:weight_trajectory}
\end{figure}

Figure~\ref{fig:weight_trajectory} visualizes the kernel weight dynamics across all benchmarks and task settings by tracking the fraction of seeds for which each kernel holds the largest weight at each iteration.

On smooth, stationary benchmarks such as Branin, the ensemble weights consistently collapse to a single dominant kernel within approximately 10--20 iterations. This \emph{winner-take-all} phenomenon is a direct consequence of the multiplicative Bayesian update rule \eqref{eq:weight-update}: when one kernel achieves marginally higher predictive likelihood at each step, the weight ratio $w_t^m / w_t^{m'}$ evolves as a product of per-step likelihood ratios, amplifying small advantages exponentially. For single-objective BO on Branin, Mat\'{e}rn-2.5 wins in 5 out of 10 seeds with an average maximum weight of 0.950, reflecting its effective balance between the infinite smoothness of RBF and the limited differentiability of Mat\'{e}rn-1.5.

The summary tables in Figure~\ref{fig:weight_trajectory} reveal that kernel preference is benchmark-dependent. On ZDT2 and Vehicle Safety (MOBO), RBF dominates universally (10/10 seeds, average max weight 1.000), while on Penicillin, Mat\'{e}rn-1.5 wins in 8/10 seeds, reflecting the less smooth objective landscape. For active learning, Mat\'{e}rn-2.5 is most frequently preferred (winning on 3 of 4 benchmarks), consistent with the observation that active learning objectives benefit from moderate smoothness assumptions for global function approximation.

Rather than harming performance, the winner-take-all behavior reflects successful automatic model selection. The ensemble efficiently identifies the most appropriate kernel for each target function without manual kernel selection or expensive cross-validation.

\subsection{Sensitivity to Scalarization Strategy}
\label{app:ablation-scalarization}

KENDO-MO reduces multi-objective conditioning to single-optimizer
conditioning via random scalarization. To assess whether its gains
depend on this specific choice, we compare the default linear scalarization
$g_{m,s}(\mathbf{x}) = \sum_k \lambda_{s,k}\, f^k_{m,s}(\mathbf{x})$
against Chebyshev scalarization
$g_{m,s}(\mathbf{x}) = \min_k \lambda_{s,k}\,(f^k_{m,s}(\mathbf{x}) - z^*_k)$,
keeping all other components (kernel ensemble, Bayesian weights, RFF sampling,
Hellinger-based acquisition) fixed.

\begin{figure}[t]
    \centering
    \includegraphics[width=0.75\textwidth]{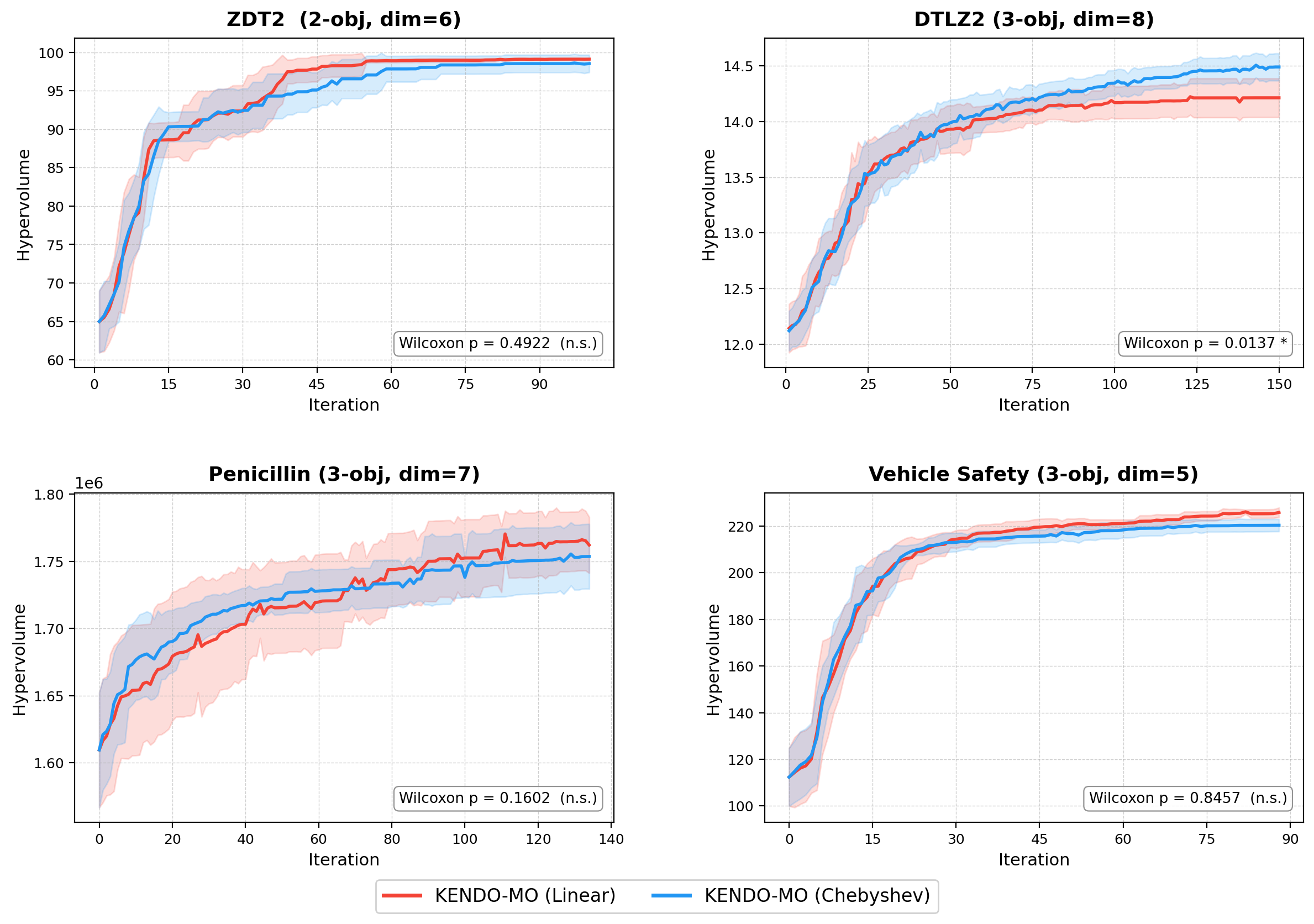}
    \caption{KENDO-MO with linear and Chebyshev scalarization on
    four MOO benchmarks. Three of four benchmarks show no statistically
    significant difference (two-sided Wilcoxon
rank-sum test on final HV); DTLZ2 favors Chebyshev
    due to its concave Pareto front
    (see Appendix~\ref{app:ablation-scalarization}).}
    \label{fig:scal-ablation}
\end{figure}

\textit{Robustness across benchmarks.}
Figure~\ref{fig:scal-ablation} reports hypervolume curves on four MOO benchmarks. A two-sided Wilcoxon rank-sum test on final hypervolume yields
no statistically significant difference on ZDT2 ($p=0.49$), Penicillin ($p=0.16$),
or Vehicle Safety ($p=0.85$). This indicates that the gains of KENDO-MO arise from
the kernel ensemble and disagreement-aware acquisition, \emph{not} from the
specific scalarization.

\begin{figure}[!htbp]
    \centering
    \includegraphics[width=0.55\textwidth]{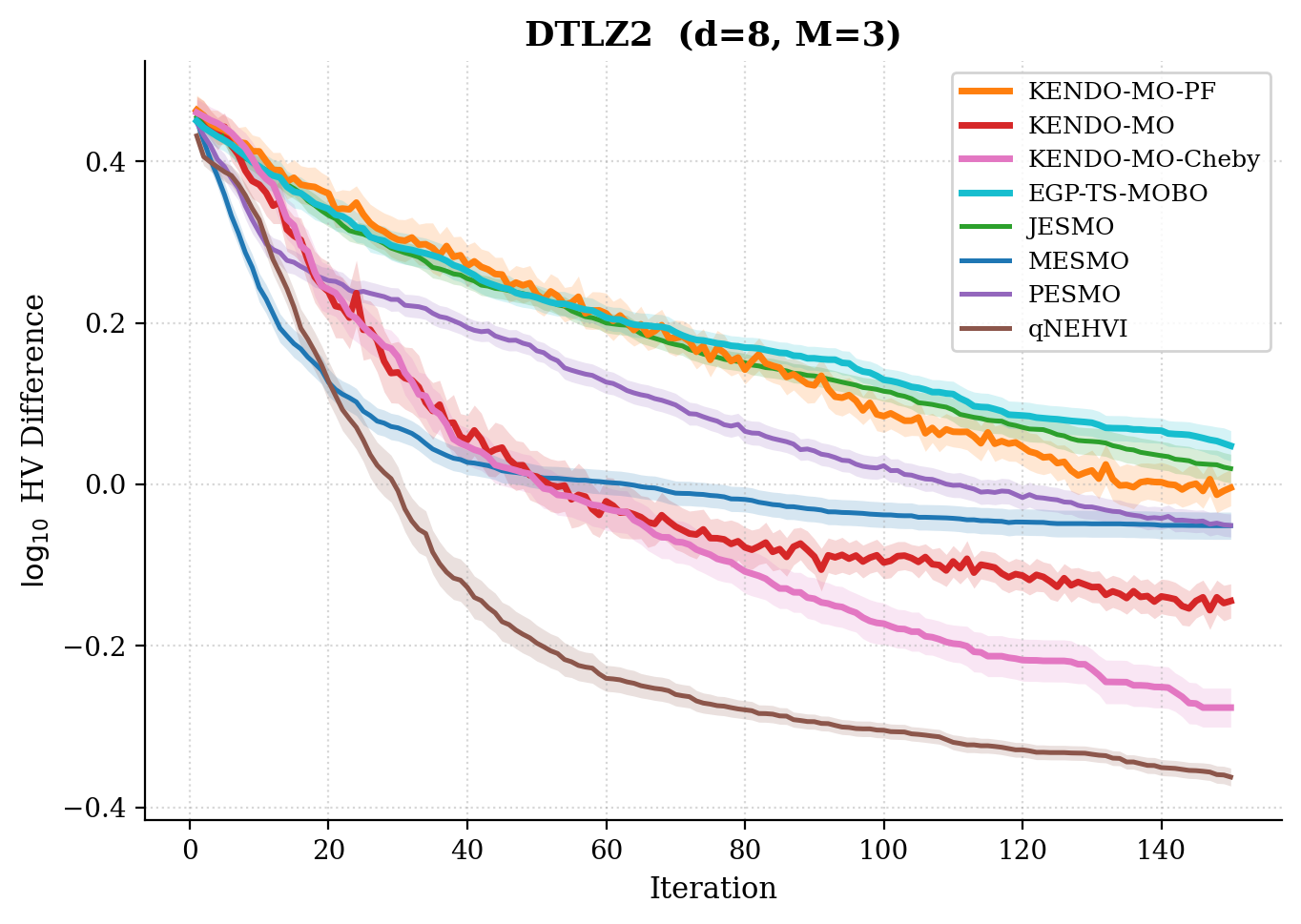}
    \caption{DTLZ2 with all multi-objective baselines plus KENDO-MO-Cheby. Both KENDO-MO variants
    outperform all entropy-search and ensemble-TS baselines; only qNEHVI
    (hypervolume-specialized) attains lower $\log_{10}$ HV difference, confirming
    that the scalarization choice is a tunable interface rather than a framework
    bottleneck.}
    \label{fig:dtlz2-cheby}
\end{figure}

\textit{The DTLZ2 case.}
On DTLZ2, Chebyshev attains a small but significant advantage ($p=0.014$). This is
consistent with a well-known property in multi-objective optimization: linear
scalarization is known to miss solutions on non-convex regions of the Pareto front,
whereas Chebyshev scalarization can recover them~\citep{knowles2006parego}. DTLZ2's
Pareto front is concave (a unit-sphere octant), so Chebyshev's edge here reflects a
geometric property of the benchmark.
To verify that the framework remains competitive on DTLZ2 once paired with the
geometry-appropriate scalarization, Figure~\ref{fig:dtlz2-cheby} plots
KENDO-MO-Cheby alongside all multi-objective baselines:
KENDO-MO-Cheby substantially outperforms all entropy-search
(JESMO/MESMO/PESMO) and ensemble-TS (EGP-TS-MOBO) baselines; only qNEHVI, a
hypervolume-specialized acquisition, achieves lower $\log_{10}$ HV difference.
The scalarization is thus a tunable interface.

Two reasons motivate linear scalarization as the default. First, the linear
objective is smooth in $\mathbf{x}$, so the $S \times M$ inner $\arg\max$ problems
on the RFF path (Algorithm~\ref{alg:KENDO-MO}, line~10) are solved efficiently with
L-BFGS; the $\min$ operator in Chebyshev is non-smooth and requires either
subgradient methods or a smoothed surrogate, inflating per-iteration cost, which conflicts with our efficiency-oriented design (Appendix~\ref{app:ablation-efficiency}). Second, linear scalarization matches the
standard ParEGO formulation~\citep{knowles2006parego}, facilitating fair comparison
with prior multi-objective BO literature.

\textit{Takeaway.}
KENDO-MO is robust to the scalarization choice, and a
practitioner may select either based on prior knowledge of the problem
geometry, e.g., Chebyshev when the Pareto front is suspected to be non-convex.

\subsection{Sensitivity to Dirichlet Concentration $\alpha$}
\label{app:ablation-dirichlet}

KENDO-MO samples scalarization weights from a symmetric Dirichlet
$\boldsymbol{\lambda}_s \sim \text{Dir}(\alpha \mathbf{1}_K)$, whose density
\begin{equation}
p(\boldsymbol{\lambda}; \alpha) = \frac{\Gamma(K\alpha)}{\Gamma(\alpha)^K}
\prod_{k=1}^K \lambda_k^{\alpha-1},
\qquad \boldsymbol{\lambda} \in \Delta^{K-1},
\end{equation}
is governed by a single concentration parameter $\alpha$ that controls how mass
is distributed on the simplex. The per-coordinate variance
$\text{Var}[\lambda_k] = (K-1)/(K^2(K\alpha+1))$ is monotonically decreasing in
$\alpha$, so $\alpha \to 0$ pushes $\boldsymbol{\lambda}$ toward the vertices
(corner-heavy, near single-objective directions), $\alpha = 1$ recovers the
uniform prior used in Section \ref{KENDO for MOBO}, and $\alpha \to \infty$ concentrates on the
centroid $(1/K, \ldots, 1/K)$ (center-heavy, near equal-weight trade-offs).
Through the inner $\arg\max$ in Algorithm~\ref{alg:KENDO-MO} (line~10),
$\alpha$ therefore directly governs the diversity of the phantom optimizers
$\{x^*_{m,s}\}$ that drive the disagreement signal in \eqref{eq:kendo-mo}.
We sweep $\alpha \in \{0.2,\, 1.0,\, 5.0\}$ on five MOO benchmarks
(LCBench, VehicleSafety, DTLZ2, Penicillin, ZDT2).

\begin{figure}[!htbp]
    \centering
    \includegraphics[height=0.70\textheight]{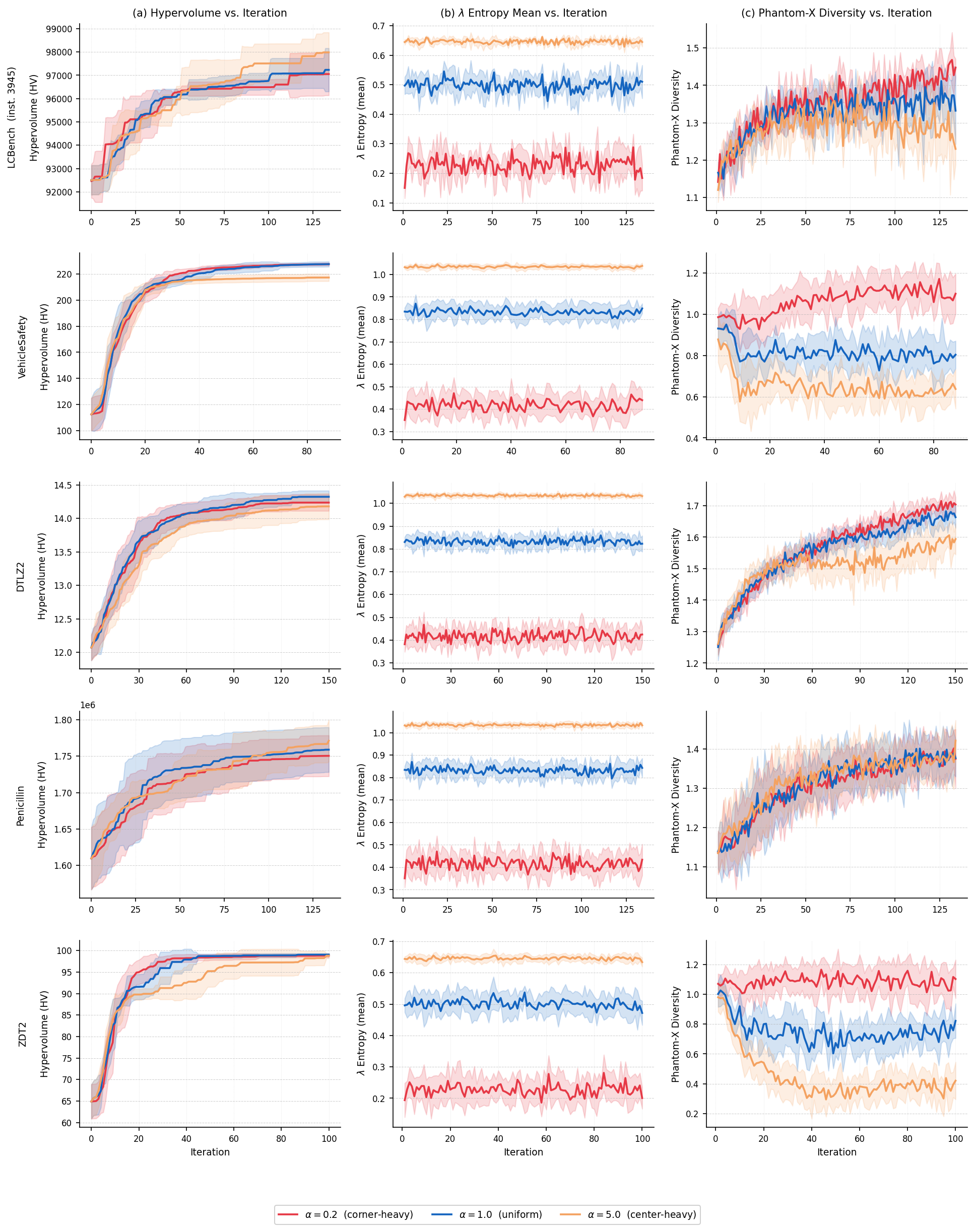}
    \caption{Sensitivity of KENDO-MO to Dirichlet concentration
    $\alpha \in \{0.2, 1.0, 5.0\}$ on five MOO benchmarks.
    \textbf{Column (a):} hypervolume vs.\ iteration.
    \textbf{Column (b):} mean Shannon entropy of sampled $\boldsymbol{\lambda}$,
    confirming that the Dirichlet sampling layer behaves as designed.
    \textbf{Column (c):} mean pairwise distance of phantom optimizers, showing
    $\alpha$ controls phantom-X diversity inversely. The default $\alpha=1.0$ is
    never significantly outperformed; optimal $\alpha$ is benchmark-dependent
    and tracks Pareto-front geometry.}
    \label{fig:dirichlet-ablation}
\end{figure}

Figure~\ref{fig:dirichlet-ablation}(b,c) verifies that $\alpha$ acts as designed.
The mean sample entropy
$\bar{H} = \mathbb{E}_{\boldsymbol{\lambda}}[-\sum_k \lambda_k \log \lambda_k]$
stratifies cleanly across the three values and approaches its ceiling $\log K$
at $\alpha=5.0$ ($\bar{H}\!\approx\!0.65$ for $K\!=\!2$ and $1.03$ for $K\!=\!3$),
while the mean pairwise distance of phantom optimizers (column c) shows the
expected inverse relationship, $\alpha=0.2$ yields the highest phantom-X
diversity and $\alpha=5.0$ the lowest. The Dirichlet sampling layer therefore
behaves exactly as the variance formula predicts.

The hypervolume curves in column (a) show that the default $\alpha=1.0$ is
robust across all five benchmarks: it is never significantly outperformed by
either extreme. The optimal $\alpha$ is, however, benchmark-dependent and tracks
Pareto-front geometry. On benchmarks with concave fronts (ZDT2, DTLZ2) and
heterogeneous landscapes (VehicleSafety), small $\alpha$ matches or exceeds the
default, consistent with the classical result that linear scalarization on
concave fronts benefits from corner-heavy directions to recover non-convex
regions~\citep{knowles2006parego}. Conversely, on real-world problems whose
trade-off region concentrates near the simplex centroid (LCBench, Penicillin),
$\alpha=5.0$ yields a small but consistent gain. Phantom-X diversity is
\emph{not} monotonically tied to hypervolume: sufficient diversity is necessary,
but excess diversity wastes budget on extreme corners that lie far from the
achievable front.

\textit{Takeaway.}
KENDO-MO's gains arise from the kernel ensemble and disagreement-aware
acquisition, not from a finely tuned Dirichlet concentration; the symmetric
prior $\alpha=1.0$ is a safe and competitive default, while practitioners with
prior knowledge of front geometry may tune $\alpha$ accordingly. Together with
the study in
Appendix~\ref{app:ablation-scalarization}, this confirms that the scalarization
layer is a tunable interface rather than a sensitive bottleneck.

\section{Additional preliminaries}

For each Gaussian process model $m$ in the ensemble, using the data $\mathcal{D}_t:=\{({\bf x}_\tau, y_\tau) \}_{\tau = 1}^t$ acquired at slot $t$, the corresponding posterior pdf $p(f(\mathbf{x})|m, \mathcal{D}_t)$ is updated via Bayes rule as \citep{rasmussen2006gaussian}:

\begin{align}
p(f({\bf x})|m, \mathcal{D}_t) = \mathcal{N}(\mu_{m,t} ({\bf x}), \sigma_{m,t}^2 ({\bf x}))
\end{align}
where
\begin{subequations}
\begin{align}
\mu_{m,t} ({\bf x}) & = \mathbf{k}_t^{\top} ({\bf x}) (\mathbf{K}_t + \sigma_{n}^2
 \mathbf{I}_t)^{-1} \mathbf{y}_t \label{eq:mean_m}\\
\sigma_{m,t}^2 ({\bf x})& = \!\kappa(\mathbf{x},\mathbf{x})\! -\! \mathbf{k}_t^{\top} ({\bf x}) (\mathbf{K}_t\! +\! \sigma_{n}^2 \mathbf{I}_t)^{-1} \mathbf{k}_t ({\bf x}) .\label{eq:variance_m}
\end{align}\label{eq:plain_gpp}
\end{subequations}
with $\mathbf{k}_t ({\bf x}) := [\kappa(\mathbf{x}_1, \mathbf{x}), \ldots, \kappa(\mathbf{x}_t,  \mathbf{x})]^\top$, $\mathbf{K}_t$ denoting the $t \times t$ kernel/covariance matrix where $[\mathbf{K}]_{i,j} = \kappa(\mathbf{x}_i, \mathbf{x}_j)$, and $\sigma_n^2$ denoting the observation variance.

Note that the mean in \eqref{eq:mean_m} provides a point estimate of $f({\bf x})$, while the variance in \eqref{eq:variance_m} quantifies the associated uncertainty.

\end{document}